\pdfoutput=1

\documentclass[11pt]{article}

\usepackage[final]{acl}

\usepackage{times}
\usepackage{latexsym}

\usepackage[T1]{fontenc}

\usepackage[utf8]{inputenc}

\usepackage{microtype}

\usepackage{inconsolata}

\usepackage{graphicx}
\usepackage{comment}
\usepackage{amssymb}
\usepackage{amsmath}
\usepackage{adjustbox}
\usepackage{multirow}
\title{A Word is Worth 4-bit: \\ Efficient Log Parsing with Binary Coded Decimal Recognition}

\author{
  Prerak Srivastava, Giulio Corallo, Sergey Rybalko \\
  SAP Labs \\
  Mougins, France \\
  \texttt{\{prerak.srivastava, giulio.corallo, sergey.rybalko\}}@sap.com
  }

\begin{document}
\maketitle
\begin{abstract}
System-generated logs are typically converted into categorical log templates through parsing. These templates are crucial for generating actionable insights in various downstream tasks.
However, existing parsers often fail to capture fine-grained template details, leading to suboptimal accuracy and reduced utility in downstream tasks requiring precise pattern identification.
We propose a character-level log parser utilizing a novel neural architecture that aggregates character embeddings. Our approach estimates a sequence of binary-coded decimals to achieve highly granular log templates extraction. 
Our low-resource character-level parser, tested on revised Loghub-2k and a manually annotated industrial dataset, matches LLM-based parsers in accuracy while outperforming semantic parsers in efficiency.
\end{abstract}

\begin{figure*}[ht]
  \centering
\includegraphics[width=0.90\linewidth]{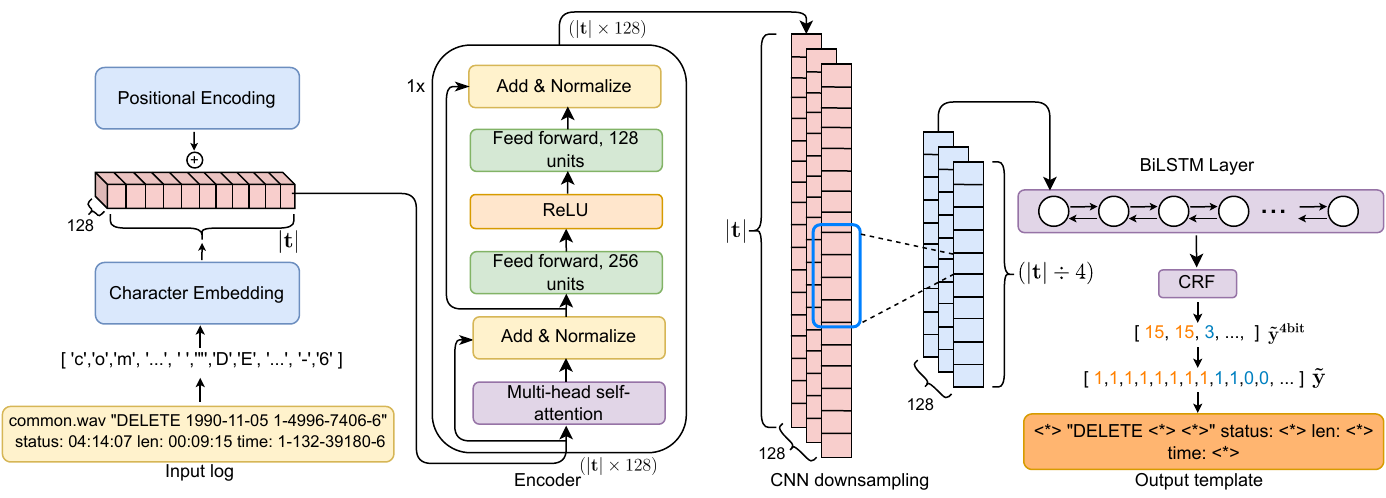}
  \caption{Diagram of the proposed neural network architecture.}
  \label{fig:diagram_architecture}
\end{figure*}
\section{Introduction}
\subsection{Existing Methods and Challenges}
Log parsing aims to extract static (log template) and variable (parameters) components from log messages, serving as a foundation for many downstream tasks such as anomaly detection \cite{steidl2022requirements}, security enhancement \cite{svacina2020vulnerability}, and audit trail \cite{wu2017using}.  As large-scale systems generate sheer amount of log data \cite{dai2020logram}, the demand for fast, automated, in-memory log parsing solutions has grown rapidly.
Traditional syntax-based parsers like Logram \cite{dai2020logram}, Lenma \cite{shima2016length}, Drain \cite{he2017drain} and Brain \cite{yu2023brain} offer computationally efficient solutions but rely heavily on pre-defined hyperparameters and manually crafted regex or grok patterns \cite{zhu2019tools}. This dependence often limits their ability to capture the deeper semantics of log data \cite{zhong2024logparser}. In contrast, semantic-based log parsers leverage deep learning for better context understanding. Semparser \cite{huo2023semparser} and Uniparser \cite{liu2022uniparser} utilize Bidirectional LSTM (BiLSTM), while NuLog \cite{nedelkoski2021self} employs a transformer-based encoder. LogPPT \cite{le2023log} goes further by utilizing pre-trained RoBERTa to frame log parsing as a token classification task, and models like VALB \cite{li2023did} and LogPTR \cite{wu2024logptr} enhance semantic understanding by precisely categorizing specific parameters.

Recently, Large Language Models (LLMs) have demonstrated remarkable versatility for downstream tasks through In-context learning (ICL) \cite{gao2023makes}, which reduces the need for fine-tuning \cite{zhao2021calibrate}. This potential is particularly promising for log parsing applications \cite{le2023log}. Advances such as LILAC \cite{jiang2024lilac} and Divlog \cite{xu2024divlog} have enhanced LLM-based log parsing accuracy through optimized few-shot example selection and improved prompt design. However, deploying LLM-based solutions presents challenges, including non-deterministic behavior across models \cite{astekin2024exploratory}, high query costs \cite{xiao2024stronger}, and significant computational requirements \cite{xia2024understanding}.

 Metrics like group accuracy \cite{zhu2019tools} tend to favor templates that appear more frequently, making them valuable for log sequence anomaly detection but often insufficient for capturing the full diversity of log templates \cite{yu2023brain,zhong2024logparser}. This bias results in incomplete comparisons and creates a gap between academic research and industry requirements \cite{petrescu2023log}. Template-level metrics such as the F1-score of Template Accuracy (FTA) \cite{khan2022guidelines} offer a more balanced perspective, particularly benefiting the evaluation of less frequent templates, like error messages.

Another critical challenge in current semantic-based log parsers is the choice of tokenization strategy. Many methods tokenize logs using space characters \citep{nedelkoski2021self}, special rules \citep{liu2022uniparser}, or pre-trained tokenizers \citep{le2023log}, leading to ambiguities in determining whether sub-tokens are static or variable components. Such ambiguities reduce the granularity of templates and cause discrepancies between annotated and predicted outputs, directly affecting performance metrics \citep{zhong2024logparser}. To improve the evaluation of template accuracy at a finer level, \citet{hashemi2024logpm} proposed the Parameter Mask Agreement (PMA) metric, focusing on character-level labeling to avoid tokenization issues. In this vein, \citet{thaler2017towards} proposed an end-to-end Deep Neural Network (DNN) approach for character-level parsing, capable of estimating parameter masks as defined by \citet{hashemi2024logpm}.

\subsection{Proposed Approach}
Building on these insights, we propose an enhanced character-level log parser inspired by the work of \citet{thaler2017towards}. Our novel architecture combines a character-level transformer, BiLSTM, and Conditional Random Field (CRF) \cite{sutton2012introduction} to achieve robust and efficient log parsing.
Our approach is evaluated using both template-level and granular-level metrics on the benchmark dataset Loghub-2k \cite{zhu2023loghub}, comparing our method against leading syntactic parsers and recent LLM-based solutions. We also evaluate our model on a manually annotated industrial dataset to demonstrate its applicability to real-world conditions. Results show that our method, with only 312k parameters, outperforms syntactic and semantic parsers and matches LILAC's performance, while being 20 times more efficient.

\section{Methodology}

\subsection{Problem definition}
 A log message $l$ can be represented as a bounded sequence of characters, denoted as $\mathbf{t} = (t_j : t \in \mathbb{T},  j= 1, 2, \dots, |\mathbf{t}|)$, where $j$  is the position index of a character within the log message. The set of all characters is represented by $\mathbb{T}$ and $|\mathbf{t}|$ indicates the total number of characters in $l$. Similar to the work of \citet{hashemi2024logpm}, for a given log message and its potential template, we aim to compute a parameter mask. This parameter mask is a binary sequence that aligns with the characters in the log message, where each binary element acts as a flag indicating whether the corresponding character is a variable or static part of the template.  The parameter mask can be described as a function $m$ that maps each character $t_{j}$ as :
 \begin{equation}
    m(t_{j}) = \begin{cases} 
0 & \text{if } t_{j} \text{ is static} \\
1 & \text{if } t_{j} \text{ is variable} . 
\end{cases}
\end{equation}
Further, we define $\mathbf{y} = (y_{1},\dots, y_{j})$ as the sequence of parameter mask, where $y_{j} = m(t_{j})$. The approach proposed by \citet{thaler2017towards} estimates $\mathbf{y}$ as a sequence of tags. To improve upon this approach, we propose replacing the final linear layer with a CRF layer, which benefits from utilizing past and future tag information to predict the current tag \cite{ma2016end}. 
 To enhance both training and inference efficiency, we propose aggregating groups of four binary labels and encoding them as a single Binary-Coded Decimal (BCD). Specifically, we define a function $m^{d}$ that maps a 4-bit sequence into a decimal number. The new sequence of decimals is represented as $\mathbf{y^{4bit}} = (d_{0}, \dots, d_{n})$, where:
\begin{equation}
    d_{n} = m^{d}(y_{4\cdot n+1}, y_{4\cdot n+2}, y_{4\cdot n+3}, y_{4\cdot n+4}),
\end{equation}
\begin{equation}
\begin{split}
    m^{d}(y_{4\cdot n+1}, y_{4\cdot n+2}, y_{4\cdot n+3}, y_{4\cdot n+4}) = 
    y_{4\cdot n+1} \cdot 2^3 \\
+  y_{4\cdot n+2} \cdot 2^2 + y_{4\cdot n+3}\cdot 2^1+ y_{4\cdot n+4}\cdot2^{0},
\end{split}
\end{equation}
and $n\in[0, (\frac{|\mathbf{t_{i}}|}{4})-1]$. The sequence $\mathbf{y}$ is padded with appropriate zeroes to meet the condition that $|\mathbf{t_{i}}|\,\% 4$. Therefore the objective of our model 
is to predict $\mathbf{y^{4bit}}$ given a sequence of characters $\mathbf{t}$ from a log line $l$, by minimizing the following loss:
\begin{equation}
    \mathcal{L}_{\boldsymbol{\theta}}(\mathbf{t},\mathbf{y^{4bit}}) =-\log p_{\boldsymbol{\theta}}(\mathbf{y^{4bit}}|\mathbf{t})\,.\tag{4}\label{eq:nll_loss}
\end{equation}

\subsection{Model architecture}
Figure \ref{fig:diagram_architecture}  illustrates our model's architecture. Without requiring pre-processing overhead, the embedding layer directly converts each input log character into a dense vector representation. Each character in the training vocabulary corresponds to a row in the embedding layer, with dimensionality set as a hyperparameter. The whitespace character is included in the vocabulary and is also used for padding the input sequence to satisfy the condition $|\mathbf{t}|\,\%\,4$. Positional encoding adds absolute positional information on each character with similar functioning as \citet{ke2020rethinking}. Next, a single-layer, 8-head bidirectional transformer encoder calculates the self-attention scores and provides mutual dependencies over past and future characters within the sequence as a hidden state. 
The hidden features generated by the encoder are aggregated in blocks of four and down-sampled using a 1D Convolutional Neural Network (CNN) with a non-overlapping kernel size and stride of 4. The down-sampled features aim to encapsulate the relevant semantic information of four characters, which are crucial for predicting $d_{n}$. The resulting sequence of down-sampled features is then passed through a single BiLSTM layer to capture dependencies in both directions. The output is fed into a CRF layer, which predicts 16 classes corresponding to the 16 possible decimal values in 4-bit BCD encoding. A more comprehensive description of the implementation details can be found in Appendix \ref{sec:appendix-implementation}. This final architecture was determined through a series of ablation studies on different blocks; see Appendix \ref{sec:appendix-ablation} for details. To train the network, we employ the negative log-likelihood loss, mentioned in Eq \ref{eq:nll_loss}. The network predicts a sequence $\mathbf{\Tilde{y}^{4bit}}$ that is mapped to binary sequence $\mathbf{\Tilde{y}}$ via a lookup table. Characters predicted as $1$ are replaced by a placeholder $<*>$ in the input log, while, the rest remain unchanged, resulting in the generation of the log template. To improve efficiency at inference, we incorporated a "parsing cache" module, inspired by the fixed-depth tree used by \citet{jiang2024lilac}. In our pipeline, this parsing cache is invoked both before querying the model and after template prediction. We refer to our method with the parsing cache as \textbf{4bitparser} and without it as \textbf{Cacheless-4bitparser} throughout the paper.

\section{Experimental Setup}

\begin{table*}[t!]
    \centering
    \footnotesize
    \begin{adjustbox}{max width=0.9\textwidth}
    \begin{tabular}{c |ccc| ccc |ccc| ccc |ccc}
    \hline
    \multirow{2}{*}{Datasets} & \multicolumn{3}{c}{Drain} & \multicolumn{3}{c}{Brain} & \multicolumn{3}{c}{LogPPT} & \multicolumn{3}{c}{LILAC} & \multicolumn{3}{c}{4bitparser} \\
    \cline{2-16}
    & PA & FTA & PMA & PA & FTA & PMA & PA & FTA & PMA & PA & FTA & PMA & PA & FTA & PMA \\
    \hline
    HDFS   & 0.35 & 0.26 & 0.97 & $\mathbf{0.50}$ & 0.33 & 0.97 & 0.44    & 0.26    & 0.88   & $\mathbf{0.50}$   & $\mathbf{0.35}$    & $\mathbf{0.98}$    & 0.44    & 0.28    & $\mathbf{0.98}$    \\
    Apache      & 0.69 & 0.50 & 0.70 & 0.69 & 0.66 & 0.70 & 0.97    & 0.61   & 0.98    & $\mathbf{1.00}$    & $\mathbf{1.00}$    & $\mathbf{1.00}$    & 0.99    & 0.83    & 0.99    \\
    OpenSSH     & 0.50 & 0.44 & 0.84 & 0.28 & 0.22 & 0.90 & 0.39    & 0.12    & 0.40    & $\mathbf{0.74}$    & $\mathbf{0.70}$    & $\mathbf{0.97}$    & 0.57    & 0.40    & 0.96    \\
    Openstack   & 0.01 & 0.01 & 0.67 & 0.11 & 0.20 & 0.95 & 0.11    & 0.16    & 0.13    & 0.40    & $\mathbf{0.76}$    & $\mathbf{0.96}$    & $\mathbf{0.44}$    & 0.73    & $\mathbf{0.96}$    \\ 
    HPC         & 0.63 & 0.36 & 0.91 & 0.66 & 0.38 & 0.82 & 0.65    & 0.48    & 0.80    & 0.88    & 0.71    & 0.97    & $\mathbf{0.99}$    & $\mathbf{0.80}$    & $\mathbf{0.99}$    \\
    Zookeeper   & 0.49 & 0.35 & 0.96 & 0.50 & 0.44 & 0.97 & 0.50    & 0.50    & 0.81    & 0.52    & 0.56    & 0.98    & $\mathbf{0.67}$    & $\mathbf{0.61}$    & $\mathbf{0.98}$    \\
    Healthapp   & 0.23 & 0.10 & 0.52 & 0.33 & 0.42 & 0.49 & 0.16    & 0.40    & 0.42    & 0.75    & 0.74    & 0.87    & $\mathbf{0.83}$    & $\mathbf{0.76}$    & $\mathbf{0.96}$    \\
    Hadoop      & 0.26 & 0.27 & 0.59 & 0.34 & 0.41 & 0.71 & 0.36    & 0.40    & 0.77    & 0.44   & 0.51    & 0.93    & $\mathbf{0.58}$    & $\mathbf{0.66}$    & $\mathbf{0.94}$    \\
    Spark       & 0.35 & 0.36 & 0.92 & 0.37 & 0.44 & 0.94 & 0.37    & 0.38    & 0.51    & $\mathbf{0.95}$    & $\mathbf{0.68}$    & $\mathbf{0.98}$    & 0.88    & 0.63    & 0.96    \\
    BGL         & 0.34 & 0.21 & 0.74 & 0.41 & 0.25 & 0.80 & 0.42    & 0.28    & 0.42    & 0.94    & 0.76    & 0.98    & $\mathbf{0.95}$    & $\mathbf{0.76}$    & $\mathbf{0.99}$    \\
    Linux       & 0.18 & 0.43 & 0.71 & 0.17 & 0.49 & 0.78 & 0.10    & 0.44    & 0.21    & 0.19    & $\mathbf{0.69}$    & $\mathbf{0.89}$    & $\mathbf{0.27}$    & 0.65    & 0.72    \\
    Mac         & 0.21 & 0.19 & 0.58 & 0.34 & 0.33 & 0.73 & 0.24    & 0.25    & 0.35    & 0.40    & 0.41    & $\mathbf{0.81}$    & $\mathbf{0.42}$    & $\mathbf{0.46}$    & 0.78    \\ 
    Thunderbird & 0.04 & 0.24 & 0.76 & 0.06 & 0.37 & 0.81 & 0.08    & 0.41    & 0.37    & 0.40    & $\mathbf{0.54}$    & 0.89    & $\mathbf{0.81}$    & 0.52    & $\mathbf{0.92}$    \\
    \hline
    \textbf{Average}     & 0.32 & 0.28 & 0.76 & 0.36 & 0.37 & 0.81 & 0.37 & 0.36 & 0.55 & 0.62 & \textbf{0.63} & \textbf{0.93} & \textbf{0.68} & 0.62 & 0.92 \\
    \hline
    \hline
    Idata       & 0.56 & \textbf{0.67} & 0.90 & 0.50 & 0.31 & 0.77 & 0.58 & 0.38 & 0.91 & 0.90 & 0.60 & 0.96 & \textbf{0.93} & 0.53 & \textbf{0.98} \\
    \hline
    \end{tabular}
    \end{adjustbox}
    \caption{Comparasion with state of the art log parsers on sub-datasets of Loghub-2k and Idata. The best results are in bold.}
    \label{tab:real_results}
\end{table*}

The network is trained on Loghub-2.0 \cite{jiang2023large} and evaluated using a refined version of Loghub-2k \cite{zhu2023loghub}, as it is a standard benchmark dataset used in the literature. The training set consists of 50k log lines from 2,349 unique templates, selected from 50 million log lines in Loghub-2.0. In line with \citet{thaler2017towards}, we prioritize diversity in the train set with a limit of 50 log lines per template. The revised Loghub-2k dataset features corrected ground truth templates following the guidelines of \citet{khan2022guidelines}, ensuring higher annotation quality. 
It includes 28k log lines from 1,139 unique templates, all excluded from the training set. Additionally, we also test on manually annotated logs from an internal cloud system referred to as \textbf{Idata}. This dataset comprises of $2400$ log lines across $35$ templates, of which $200$ log lines are used to fine-tune the trained network and adjust baseline models, we refer this subset as \textbf{Idata-FT}. Further details regarding the log types and annotation strategies for these datasets are provided in Appendix \ref{sec:appendix-dataset}. To gauge the performance between different log parsers, we use three metrics: Parsing Accuracy (PA) \cite{dai2020logram}, FTA \cite{khan2022guidelines} and PMA \cite{hashemi2024logpm}. PMA is computed by generating a ground truth parameter mask for each log template in both test sets, as described by \citet{hashemi2024logpm}. 
 For our experiments, we train 4bitparser using a defined set of hyperparameters (Appendix \ref{sec:appendix-implementation}).

We compare 4bitparser with syntactic parsers such as Drain and Brain, as well as semantic parser-based models like LogPPT and LILAC, with LILAC being specifically LLM-based. The syntactic parsers are evaluated using the pre-defined benchmark hyperparameters on the Loghub-2k test set, while hyperparameters for Idata are fine-tuned separately. LogPPT is trained on data similar to 4bitparser, including its training set and Idata-FT for the respective tests. For the LILAC baseline, we sampled 128 candidate examples and selected three demonstrations for the GPT-3.5-turbo language model, with the candidate set for demonstration sampled separately for both test sets following the approach of \citet{xu2024divlog}.

\begin{figure}[t!]
  \centering
  \includegraphics[width=4.5cm]{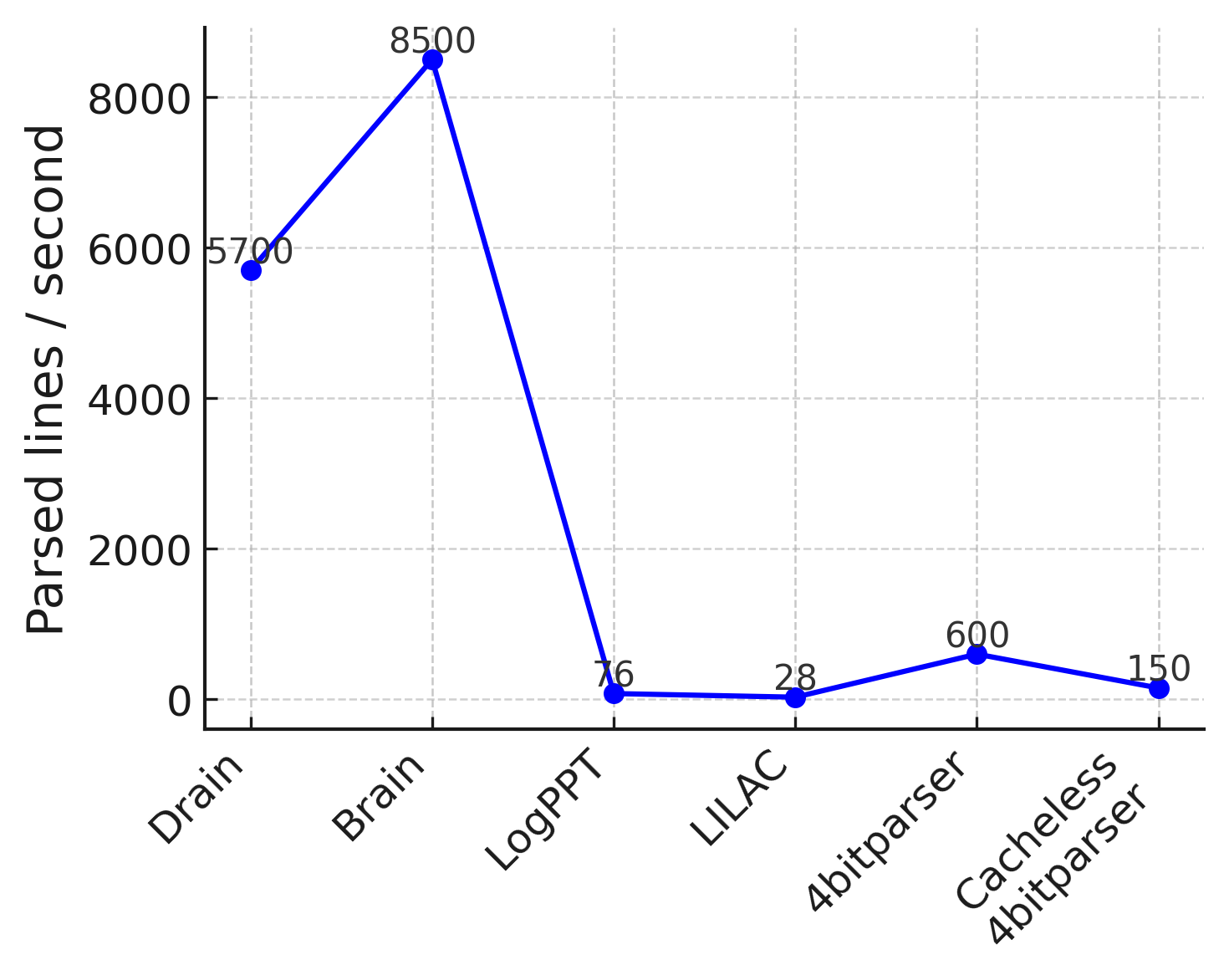}
  \caption{Throughput of log parsers on Loghub-2k.}
  \label{fig:Throughput}
\end{figure}

\section{Experimental Results}

\subsection{Effectiveness of 4bitparser}
The results, presented in Table \ref{tab:real_results}, report performance across the Loghub-2k sub-datasets and Idata, with no pre-processing applied to the extracted or ground truth templates. On average, 4bitparser identifies nearly twice as many templates and offers 10\% greater granularity than the Syntactic parser and LogPPT. While there is minimal difference between LILAC and 4bitparser in template granularity (PMA) and exact template extraction (FTA), 4bitparser identifies 6\% more frequent templates, leading to an improvement in PA. Moreover, 4bitparser outperforms LILAC on four sub-datasets and shows comparable or slightly varied results on the rest. This is remarkable given that our model is very lightweight, with only 314k parameters.
On Idata, Drain identifies more templates than the other parsers, mainly  because this test set includes log templates with identical token counts across log parameters, making them well-suited for Drain's heuristics. However, 4bitparser excels at identifying frequent templates, and even its non-exact matches are highly accurate at the character level, achieving the highest PMA score of $0.98$. 
\subsection{Efficiency of 4bitparser}
 To evaluate the efficiency of each log parser, we measured throughput in terms of log lines parsed per second on a set of 28k logs from the Loghub-2k test set. For this evaluation, all semantic parsers along with two of our proposed methods, were executed on a 16GB V100 GPU. Figure \ref{fig:Throughput} shows that Cacheless-4bitparser is five times more efficient than LILAC, which already employs a parsing cache, and twice as efficient as LogPPT, highlighting the computational overhead of language model-based log parsers. Additionally, 4bitparser, when using the parsing cache, is 20 times faster than LILAC and 8 times efficient than LogPPT, significantly narrowing the efficiency gap between language model-based and syntactic parsers.
 
\section{Conclusion}
This study presents a novel character-level log parser that extracts templates by predicting a sequence of binary-coded decimals. A parsing cache module and efficient 4-gram feature aggregation via CNN enhance its efficiency. Results on benchmark and industrial datasets show the method outperforms syntactic parsers and rivals LLM-based parsers for accurate and granular template extraction. Future work aims to develop a vocabulary-free model for broader applicability, integrate with downstream tasks, and explore unsupervised pre-training on large-scale log data.

\section{Limitations}
The character-level log parser in this study operates on character embeddings fixed using the Loghub-2.0 training data. While the trained model successfully captured most characters needed for fine-tuning and testing with our industrial dataset, Idata, a vocabulary-free approach could improve deployment efficiency. Our evaluation primarily focuses on accurate template extraction, but it can be extended to metrics like group accuracy \cite{zhu2019tools}, which are crucial for log sequence anomaly detection \cite{yu2023brain}.
To pre-train our supervised learning model, we limited the dataset to 50k logs with sufficient template diversity. However, we believe that incorporating more logs, a wider variety of templates, and longer training could improve generalization on new datasets after fine-tuning on smaller training sets. This may also require adjustment in the current hyper-parameters used in this work. Additionally, we did not address the continuous learning loop \cite{van2019three}, which is crucial for real-world deployment.

\bibliography{custom}

\appendix

\section{Datasets Statistics}
\label{sec:appendix-dataset}
\subsection{Loghub}

An extensive collection of logs generated by 16 different software systems is released as Loghub by \citet{zhu2023loghub}. This dataset accounts for 490 Million log lines amounts to over 77GB in total. Nonetheless, it provides annotated ground truth for log parsing of only 2000 random log lines for each sub-dataset, named as Loghub-2k. It has been used to benchmark many different log parsing systems \citep{zhu2019tools,dai2020logram}. With the raw Loghub logs, \citet{jiang2023large} builds a large scale annotated dataset for log parsing. It consists of 50 Million log lines with 3488 unique template for 14 different software systems. Log templates are annotated adhering to a rigurous framework which considers heuristic rules given by \citet{khan2022guidelines} and parameter categories proposed by \citet{li2023did}. Similarly, Loghub-2k is also revised as per  \citet{khan2022guidelines} and is available on GitHub \footnote{\url{https://github.com/logpai/loghub-2.0/tree/main/2k\_dataset}}. We trained our method using 50k log lines from 2,349 unique templates in Loghub-2.0 and tested it on the revised Loghub-2k. Table \ref{tab:Appendix_table_1} provides detailed statistics on the datasets used. No pre-processing was applied to the training or test sets; raw annotated templates were used directly.

\subsection{Idata}
We extracted 100k log lines from an internal cloud system and sampled 6,000 for annotation. Two annotators independently labeled templates for 3,000 log lines each, resulting in 36 unique templates. Of these, 35 templates reached consensus, covering 2,400 log lines. One contested template accounted for the remaining 3,600 log lines, skews the dataset. Since accurately identifying this single template could inflate Parsing Accuracy to over 50\%, we excluded it, and only the remaining annotated log lines were used for this study.

\section{Implementation Details}
\label{sec:appendix-implementation}
\begin{table}[t!]
    \hspace*{-0.4cm}
    \footnotesize
    \begin{tabular}{c|c|c|c}
\hline
\multirow{2}{*}{Datasets} & \#Templates & \#Templates & \#No of logs \\
& (Loghub-2k) & (Loghub-2.0) & (Loghub-2.0) \\
\hline
Hadoop & 114 & 236& 179,993\\
HDFS & 14 & 46& 11,167,740\\
Openstack & 43& 48& 207,632\\
Spark  &  36& 236& 16,075,117\\
Zookeeper & 50& 89& 74,273\\
BGL & 120& 320 & 4,631,261\\
HPC & 46& 74& 429,987\\
Thunderbird & 149& 1241& 16,601,745\\
Linux &118 & 338& 23,921\\
Mac & 341& 626& 100,314\\
Apache & 6& 29& 51,977\\
OpenSSH & 27& 38& 638,946\\
HealthApp & 75& 156& 212,394\\
Proxifier & 8& 11& 21,320\\

\hline
\textbf{Total} & 1,147 & 3,488& 50,416,620\\
\hline
\end{tabular}
    \caption{ \raggedright Statistics on Loghub-2.0 and Loghub-2k}
    
\label{tab:Appendix_table_1}
\end{table}
\begin{table}[ht]
\centering
\small
\begin{tabular}{ll}
\hline
\textbf{Layer} & \textbf{Details} \\
\hline
\texttt{char\_embed} & Embedding(100, 128) \\
\texttt{attn\_self\_1} & ResidualAttentionBlock \\
 & - MultiheadAttn(128, 8 heads) \\
 & - MLP(128$\rightarrow$256$\rightarrow$128) \\
\texttt{pos\_enc} & PosEnc(dropout=0.1) \\
\texttt{cnn\_1d} & Conv1d(128, 128, k=4, s=4) \\
\texttt{lstm\_2} & BiLSTM(128, 64) \\
\texttt{ln\_3} & LayerNorm(128) \\
\texttt{crf\_layer} & CRF(Linear(128$\rightarrow$16)) \\
\texttt{dropout} & Dropout(0.4) \\
\texttt{relu} & ReLU() \\
\hline
\end{tabular}
\caption{Architecture summary of 4bitparser.}
\label{tab:architecture}
\end{table}
The character embedding dimension is set to 128. The encoder employs 8-head self-attention, and the CNN is configured with an output filter size of 128, a kernel size of 4, and a stride of 4. Following the CNN, a single-layer BiLSTM is used, with a hidden size of 64 × 2. To prevent overfitting, a dropout rate of 0.4 is applied after both the CNN and BiLSTM, with skip connections integrated at both the encoder and BiLSTM layers to mitigate vanishing gradient issues. Our model architecture summary is shown in Table \ref{tab:architecture}. The model, implemented using PyTorch Lightning \cite{falcon2019pytorch}, contains 312k parameters (4.2 MB) and is trained on a single 16GB V100 GPU or 10 epochs. We use a batch size of 16, a weight decay of $1e^{-4}$, and a learning rate of $1e^{-3}$.
We employed the Adam optimizer \citep{KingBa15}, using PyTorch’s default hyperparameters, except for the learning rate.
The code for reproducing our experiments on the public datasets will be released on GitHub.
\section{Other Tried Architectures}
\label{sec:appendix-ablation}
We tried replacing the encoder with the BiLSTM block similar to  \citet{thaler2017towards}, but this resulted in increased inference and training time with a minor drop in performance, in line with \citet{yan2019tener}. We also tried kernel sizes of $2,3,4\,$ and $5$ at the down-sampling block. Apart from $4$ other kernel sizes reported drop in performance with kernel size of $<4$ lead to increase inference time. We find that 4-gram features efficiently represent logs, as static or variable sub-tokens typically average four characters. Combination of BiLSTM and CRF is inspired by the VALB log parser \cite{li2023did}, which uses word embeddings instead our method operates on fixed-length n-grams. Removing the BiLSTM layer and replacing the CRF with a linear transform resulted in poorer performance.

\end{document}